\documentclass[10pt,a5paper,smallheadings,liststotoc,headsepline,DIVcalc,BCOR12mm,tablecaptionabove,fleqn,runningheads,final]{fbvproc}

\usepackage{etex}
\usepackage{makeidx}
\usepackage[english]{babel}
\usepackage{graphicx}
\usepackage[utf8]{inputenc}
\usepackage{tabularx}
\usepackage{amsfonts}
\usepackage[mathscr]{eucal}
\usepackage{listings}
\usepackage[font=footnotesize,labelfont=bf,format=hang]{caption} 
\usepackage{color}
\usepackage[table]{xcolor}
\usepackage[standard,thmmarks]{ntheorem}
\usepackage{url}
\usepackage[amssymb]{SIunits}
\usepackage{bibunits}
\usepackage{amsmath}
\usepackage[footnotesize,bf,hang]{subfigure} 
\usepackage{cite}
\usepackage{nicefrac}
\usepackage{dsfont}
\usepackage{enumerate}
\usepackage{wrapfig}
\usepackage{emptypage}
\usepackage{xspace}
\usepackage{textcomp}
\usepackage{tikz}
\usetikzlibrary{matrix,arrows.meta}
\usepackage{wasysym}

\usepackage[sc]{mathpazo}    
\usepackage[scaled]{helvet}  

\DeclareFontEncoding{LGR}{}{}
\DeclareTextSymbol{\~}{LGR}{126}

\newenvironment{itemize*}%
  {\begin{itemize}%
    \setlength{\itemsep}{1pt}%
    \setlength{\parskip}{1pt}}%
  {\end{itemize}}

\newenvironment{enumerate*}%
  {\begin{enumerate}%
    \setlength{\itemsep}{1pt}%
    \setlength{\parskip}{1pt}}%
  {\end{enumerate}}

\newenvironment{description*}%
  {\begin{description}%
    \setlength{\itemsep}{1pt}%
    \setlength{\parskip}{3pt}}%
  {\end{description}}

\begin{document}

	\selectlanguage{english}

	\mainmatter

%
%

\title{An Image Processing Pipeline for Automated Packaging Structure Recognition}

\titlerunning{Automated Packaging Structure Recognition}

\author{Laura D\"orr, Felix Brandt, Martin Pouls and Alexander Naumann}

\authorrunning{L. D\"orr, F. Brandt, M. Pouls and A. Naumann}

\tocauthor{L. D\"orr, F. Brandt, M. Pouls and A. Naumann}

\institute{%
FZI Research Center for Information Technology,\\
Haid-und-Neu Str. 10-14, 76131 Karlsruhe, Germany \\
\{doerr,brandt,pouls,anaumann\}@fzi.de}

\abstract{Dispatching and receiving logistics goods, as well as transportation itself, involve a high amount of manual efforts. The transported goods, including their packaging and labeling, need to be double-checked, verified or recognized at many supply chain network points. 
These processes hold automation potentials, which we aim to exploit using computer vision techniques. 
More precisely, we propose a cognitive system for the fully automated recognition of packaging structures for standardized logistics shipments based on single RGB images. 
Our contribution contains descriptions of a suitable system design and its evaluation on relevant real-world data. 
Further, we discuss our algorithmic choices.}


\maketitle 

\section{Introduction}

\begin{figure}[tb]
	\centering
	\includegraphics[width=0.4\textwidth,trim={0cm 0cm 0cm 3cm},clip=true]{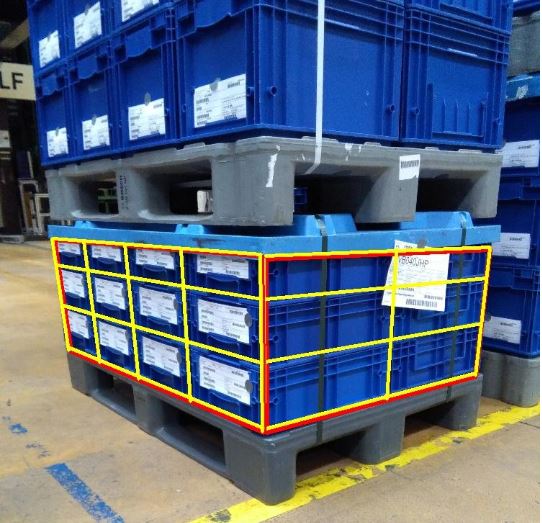}
	\caption{Illustration of Packaging Structure Recognition. Transport unit side faces are illustrated in red, yellow lines indicate packaging unit rows and columns.}
	\label{fig:psr_illustration}
\end{figure}

In logistics supply chains, goods are transported along many different network points and require to be manually checked at each of these points.
Such manual inspections often include not only unit identity but also completeness or packaging instruction compliance.
In an aim to enable further automation of such inspection processes, we designed a system for automated image-based packaging structure recognition.
Hereby, we define packaging structure recognition as the task of recognizing and analyzing logistics transport units and their building structure, allowing for inference of packaging types, number and arrangement.
Fig. \ref{fig:psr_illustration} illustrates this task.

While numerous related image-based systems have been introduced by both dedicated start-ups and experienced vision and logistics companies, we are not aware of alternative solutions to the task of 3D packaging structure recognition for standardized logistics shipments from a single RGB image. 
The tackled tasks often include the detection of single packages or multiple package shipments and their dimensions. 
In many cases, individual packages or objects are recognized and counted, or logistics transport labels are found and read.
For instance, a system by logivations \cite{Solution:Logivations} tackles a similar use-case of automated goods receive by detection and reading of logistics transport labels. 
Further, the solution is able to measure logistics units and count visible object instances.
A method proposed by Fraunhofer IML \cite{Solution:Fraunhofer_Ladungstraeger} tackles the related problem of empties counting and tracking.
Apart from solving slightly different tasks, many comparable systems use supplementary image and data acquisition means, e.g. multiple camera systems or additional sensors such as laser scanners or infrared technologies (e.g. \cite{Solution:vitronic}, \cite{Solution:cognex}).
Aside from image based methods, the usage of non-visual sensors and information, like barcodes or RFID tags, is applicable to the problem at hand. 
However, such methods are often more expensive and still error-prone, as sensor ranges are limited and hardware requirements are enormous. 
Some of the obstacles regarding RFID technologies in logistics are discussed in \cite{lim2013rfid}.
At the same time, the hardware requirements for a image-based system like ours are minimal as we make no special assumptions regarding camera hardware. 

We propose a solution for the task of packaging structure recognition based on single RGB images of applicable transport units, meeting certain necessary restrictions.
The algorithm presented in this work was previously introduced and evaluated in \cite{doerr2020psr}. 
Further, the logistics context and the use-cases we focus on are explained thoroughly in the before-mentioned publication.
In this paper, we supplement our work by discussing the algorithmic choices and conducting further experiments and evaluations.
More precisely, we analyze the task and define a series of sub-steps which, when combined together, are able to solve the task.
For each of these tasks, we discuss input and expected output, requirements and evaluate our algorithmic approaches.
We give reasons for the algorithmic choices we made, in some cases by evaluating different options on our own real-world use-case-specific data set.

\section{Problem and Setting}

In this section, we discuss the problem of packaging structure recognition more detailed and introduce clarify some logistics terms used throughout this work.
Further, the setting in which the system was designed and tested is described, and necessary restrictions are explained.

\subsection{Terms and Definitions}
\label{sec:definitions}

\textbf{Packaging Unit.}
Packaging units are any containers used to transport goods along a logistics supply chain.
Though made of various materials, these containers are often highly standardized (e.g. small load carrier system (KLT) \cite{VDA:KLT}).

\textbf{Base Pallet.}
This term is used to describe the base unit on which logistics goods can be stacked for transport. 
A wide range of mostly standardized pallets exist (e.g. EPAL Euro Pallet \cite{EPAL:pallet}).

\textbf{Transport Unit.}
A logistics transport unit refers to fully-packed, shipping ready assortment of goods.
Usually, such a unit consists of one base pallet, one or multiple packaging units and additional optional components, for instance lids, security straps or transparent foils.
When speaking of uniformly packed transport units, we refer to transport units containing only one single type of packaging units of identical size.
By regularly packed transport units, we mean units with a fully regular packaging pattern, i.e. all rows, columns and layers of packages contain the same number of packaging units.

\textbf{Transport Unit Side Face.}
We use the term transport unit side face to refer to that part of a transport unit which is visible when looking at it frontally from an arbitrary side with horizontal visual axis.
Each logistics transport unit has exactly four such side faces. 
An occlusion-free image covering a whole transport unit can at most show two neighboring side faces of the transport unit.

\subsection{Problem Formulation}
The problem of packaging structure recognition as tackled in this paper is the task of localizing and inferring the packaging structure of one or multiple stacked transport units in a single RGB image.
Hereby, the packaging structure consists of the number and type of packaging units, the arrangement of these units and, optionally, the type of base pallet present.

\subsection{Setting and Restrictions}
\label{sec:setting}
The task of packaging structure recognition as described above is not always solvable without making further assumptions on logistics components and imagery.
Thus, we try to define a setting and reasonable restrictions to achieve feasibility.

\textbf{Packaging Restrictions.}
First of all, only regularly and uniformly packed transport units are considered.
This is necessary as the packaging structure of non-regularly packed transport units can in general not be inferred by observing a single image of that unit.
Further, restricting the method to such units allows for improved robustness as not every individual packaging unit needs to be detected and identified.
Instead, one can assume the rows and columns of each transport unit side to have the same number and types of packages, which the proposed algorithm does.

\textbf{Imaging Restrictions.}
Additional restriction regarding image acquisition and contents.
All images need to be taken in an upright orientation in such a way that vertical real-world structures (such as transport unit edges) are roughly parallel to vertical image boundaries.
Further, relevant transport units within the image are shown in their full extent and not occluded by any persons or objects.
Additionally, we require transport units to be photographed in such an angle, that two of their side faces are clearly (and evenly well) visible. 

\textbf{Material Restrictions.}
For the time being, we limit our setting to a set of defined logistics components, i.e. packaging units and base ballets.
As part of the algorithm relies on learning-based methods, we can only assume generalization to what is contained in the training data.
Relevant packaging units in our setting are KLT packages and so-called tray packages, as is described more detailed in section \ref{sec:data}.

\section{Method Overview}

\begin{figure}[htb]
	\centering
	\includegraphics[height=0.17\textheight]{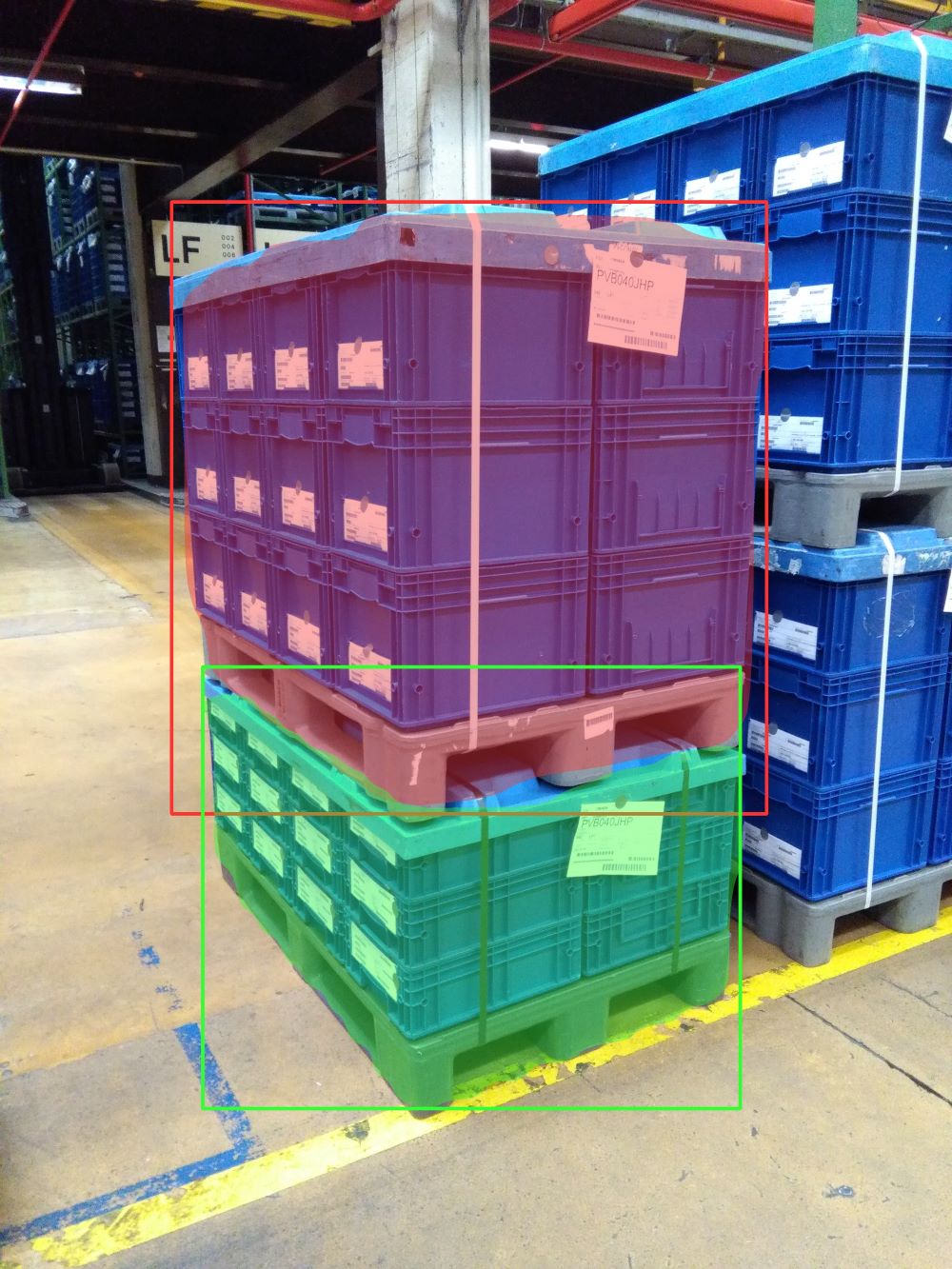}
	\includegraphics[height=0.17\textheight]{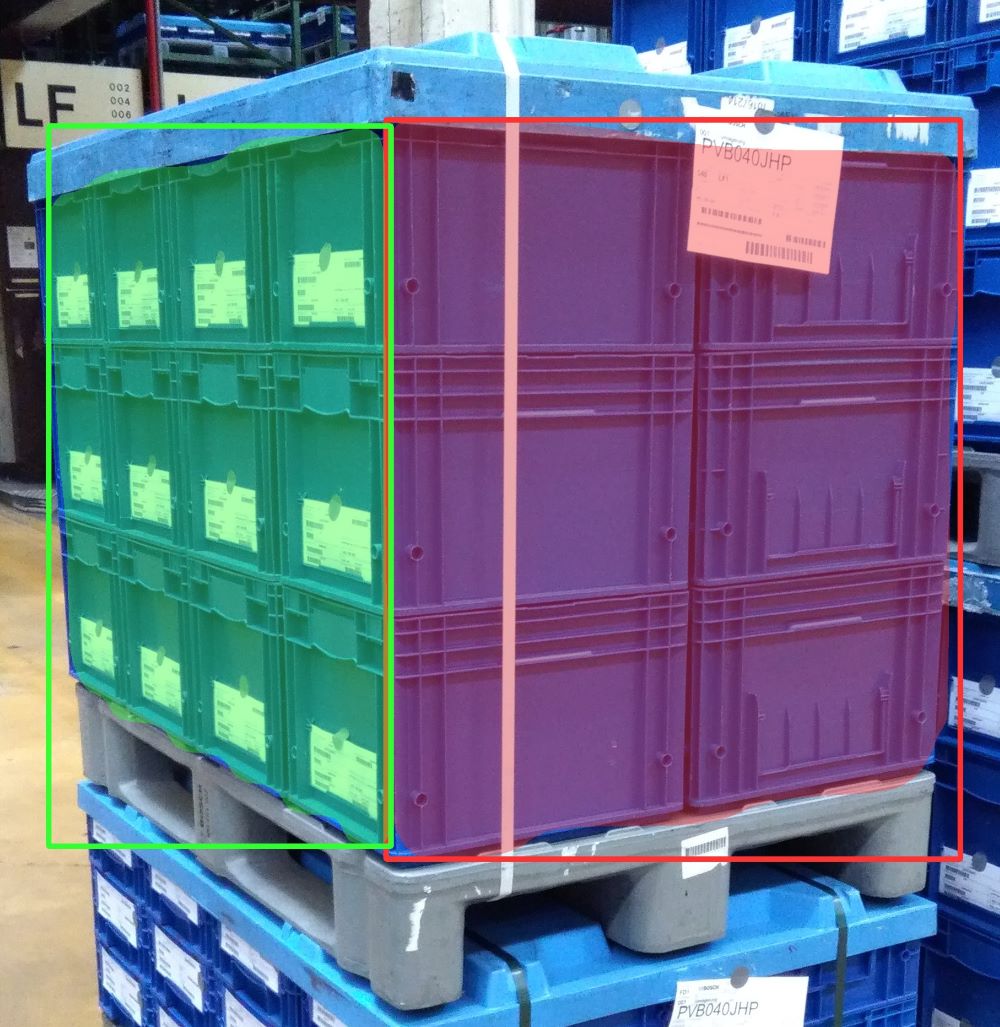}
	\includegraphics[height=0.17\textheight]{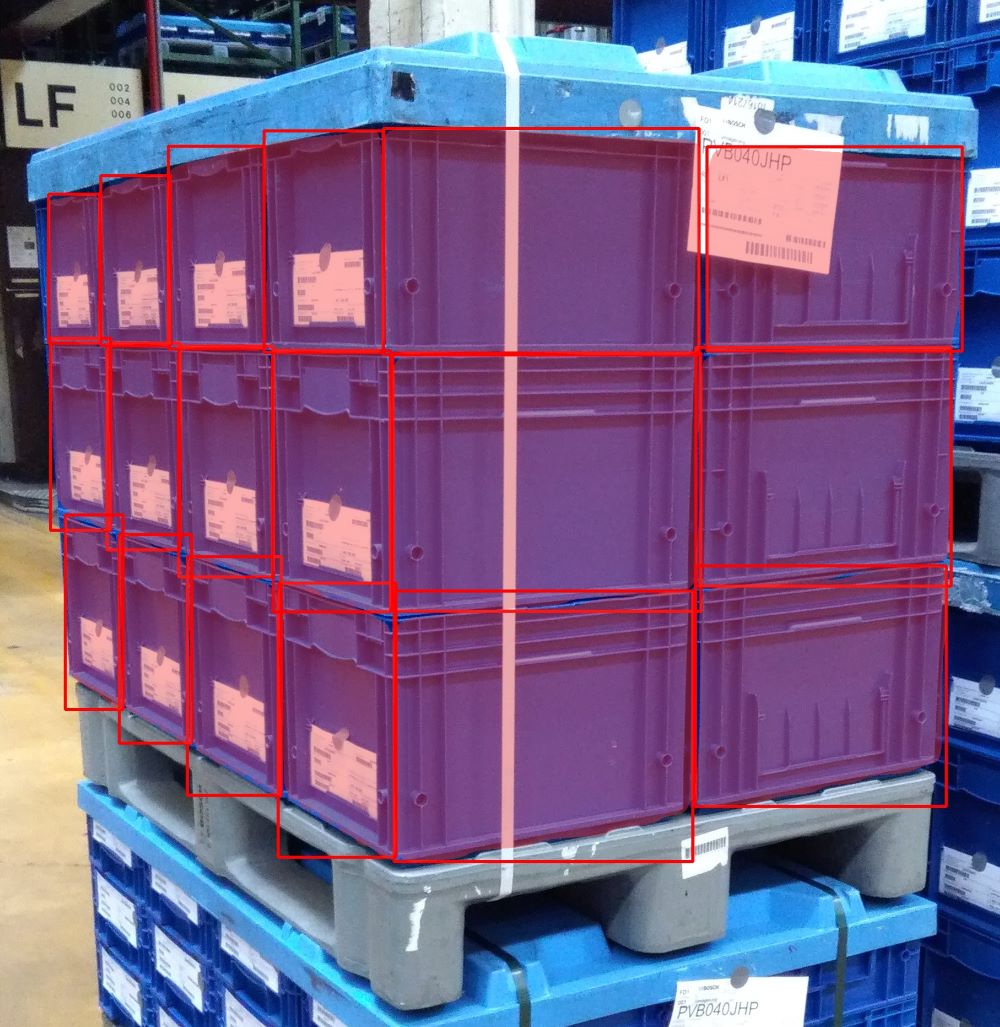}
	\includegraphics[height=0.17\textheight]{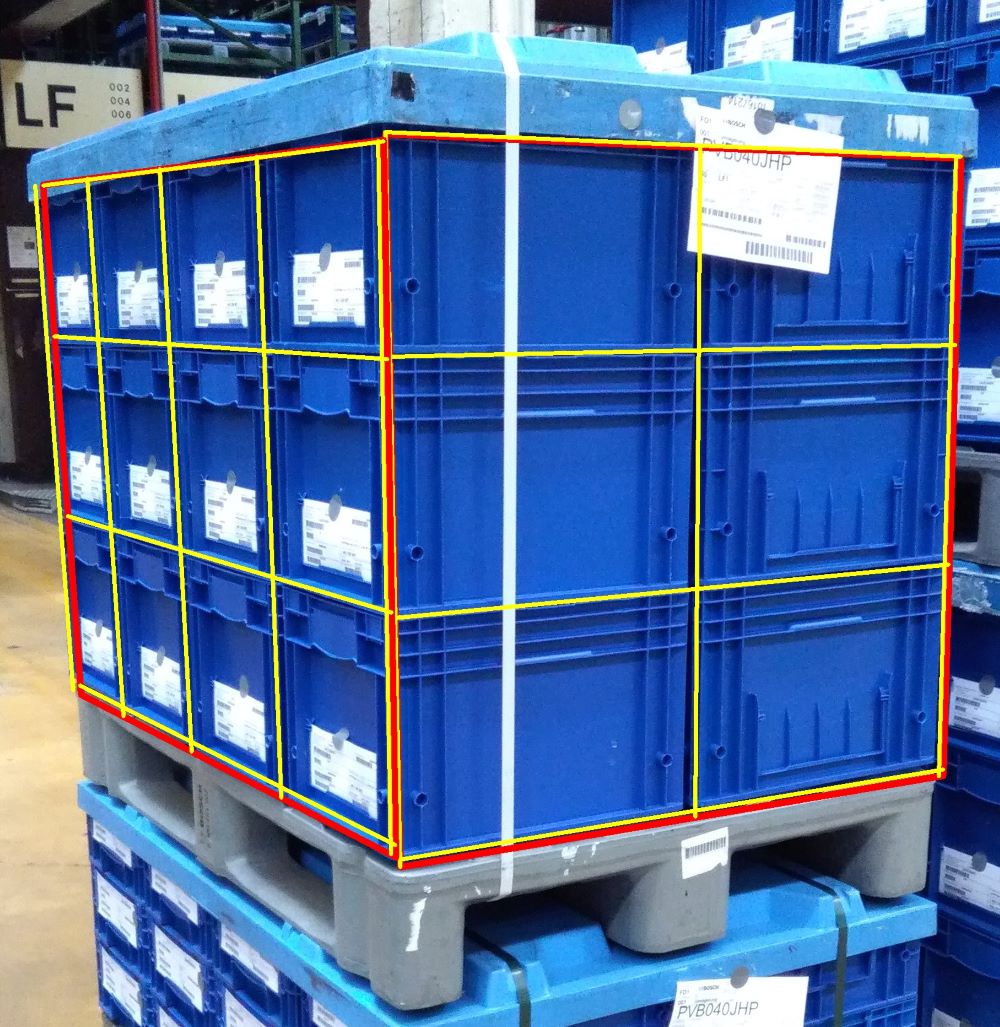} \\
	(a) \hspace{0.17\textwidth} (b) \hspace{0.20\textwidth} (c) \hspace{0.20\textwidth} (d) \hspace{0.02\textwidth}
	\caption{Method Overview. (a) Step 1: Transport unit identification. (b) Step 2: Transport unit side face segmentation. (c) Step 3: Packaging unit identification and localization. (d) Step 4: Information consolidation, and output visualization. }
	\label{fig:method_overview}
\end{figure}

This section contains a detailed description of the algorithm's coarse structure, 
i.e. we explain the series of independent tasks which build our image processing pipeline for packaging structure recognition.
The four subsequent steps are illustrated in Fig. \ref{fig:method_overview}.

\textbf{Step 1: Transport unit identification and localization.}
The first step in our pipeline is to identify and localize all relevant logistics transport units in the input image.
Relevant are such transport units which are visible in their full extend and without any occlusions.
The expected output of this task is the number and locations of all transport units within the image.
Location information consists not only of the bounding box describing the image part fully covering the transport unit, but also a pixel-based instance mask, which provides valuable information for subsequent steps.

\textbf{Step 2: Transport unit side face segmentation.}
The input of the pipeline's second processing step is a crop of the original image, containing exactly one, fully visible transport unit and the corresponding pixel mask. 
This step is performed for each transport unit found by the previous step.
The expected output of this step are two segmentation masks for the transport unit's two visible side faces (see Section \ref{sec:setting}).
Note that bounding boxes are again not sufficient, but more detailed, pixel-based information is required. 
The segmentation mask can be encoded as the coordinates of the four pixels showing the transport unit side face's four corners.
As a transport unit side face can be described by a rectangle in 3D real space, it can be exactly localized by four pixel coordinates in our image, when assuming a distortion-free projective transform is underlying the image acquisition process.

\textbf{Step 3: Packaging unit identification and localization.}
In this step, the packaging units for each transport unit side face are localized and classified.
The task's input is the cropped handling unit image (same as input to step 2) and the corresponding handling unit side face information (output of step 2).
The expected output is pixel-based information of the packaging unit's contained in each transport unit side.
As in the previous step, this information can be encoded as the coordinates of four pixels for each packaging unit found within the image.

\textbf{Step 4: Information consolidation.}
The last step used the information derived in the three previous steps to compose the desired output: the packaging structure of each transport unit contained and fully visible within the image.
The most essential part of this task is the packaging number calculation for each transport unit side face.
Here, the average width and height of each packaging unit is computed, considering the provided pixel-based segmentation information to account for varying object sizes due to perspective distortions.

\section{Experiments and Implementation}

\subsection{Data Set}
\label{sec:data}

\begin{figure*}[htb]
	\centering
	\includegraphics[height=0.15\textheight,trim={0cm 22cm 0cm 15cm},clip=true]{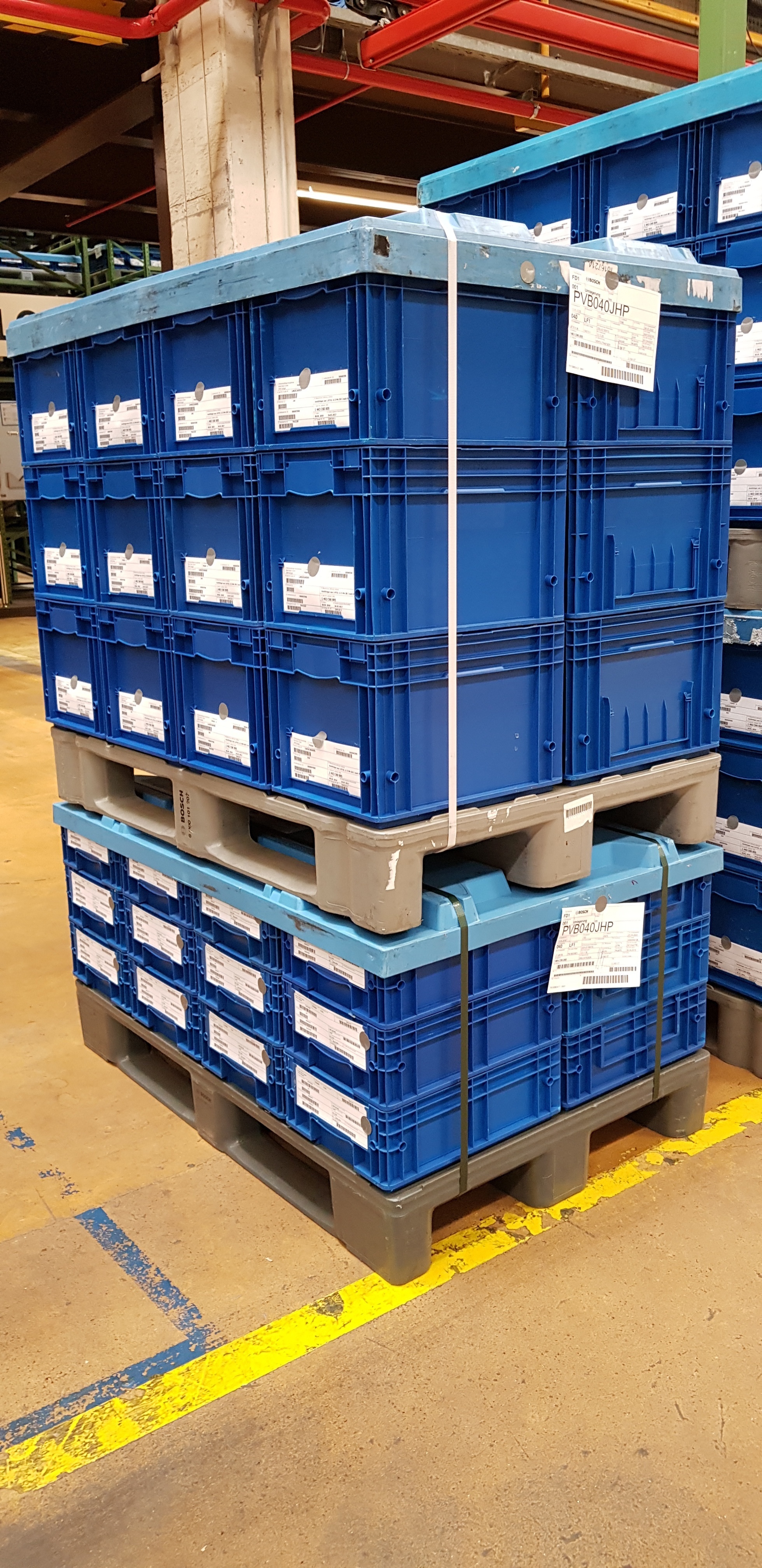}
	\includegraphics[height=0.15\textheight,trim={0 12cm 0 18cm},clip=true]{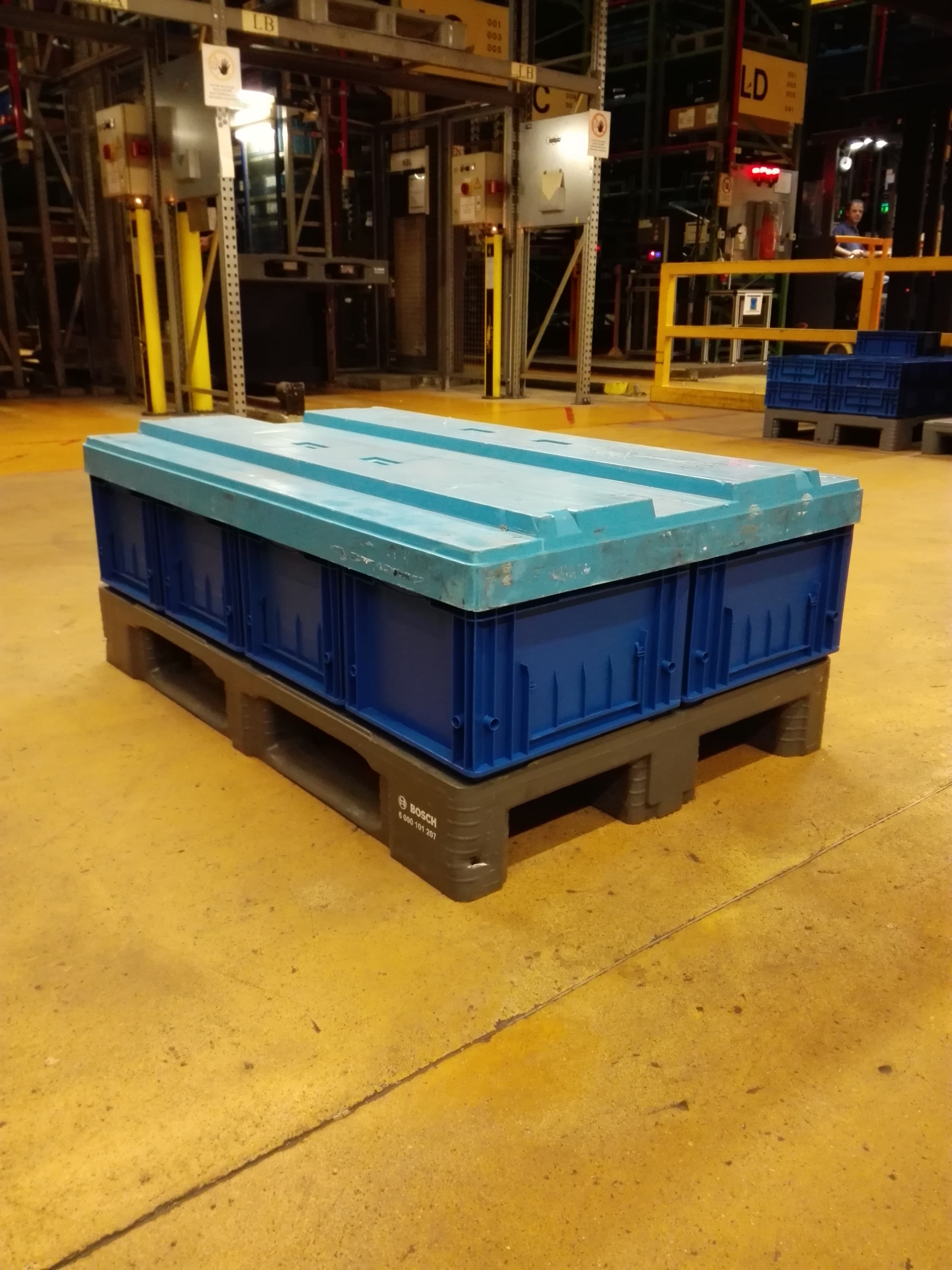}
	\includegraphics[height=0.15\textheight,trim={3cm 5cm 0 28cm},clip=true]{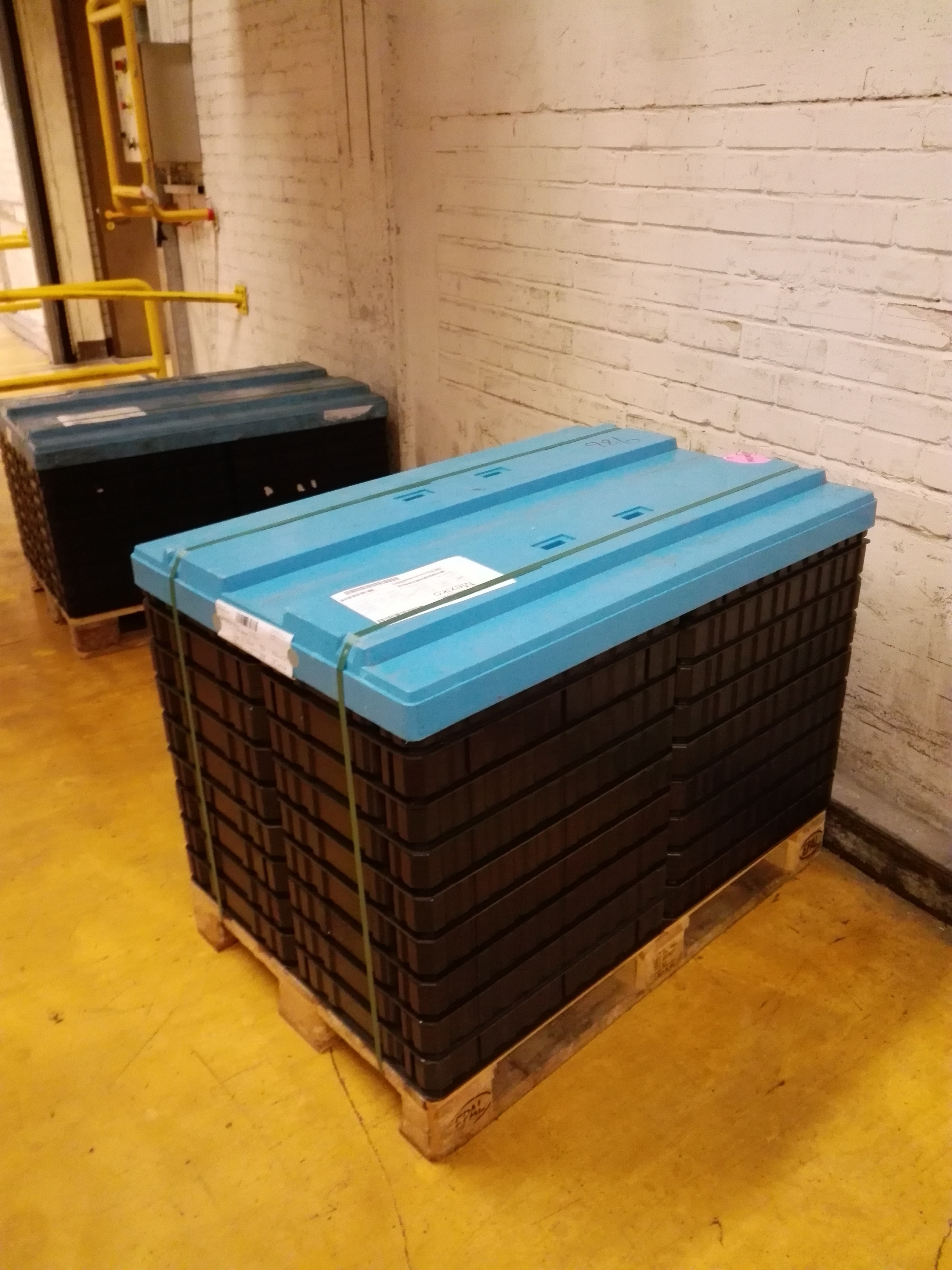}
	\includegraphics[height=0.15\textheight,trim={0 15cm 10cm 26cm},clip=true]{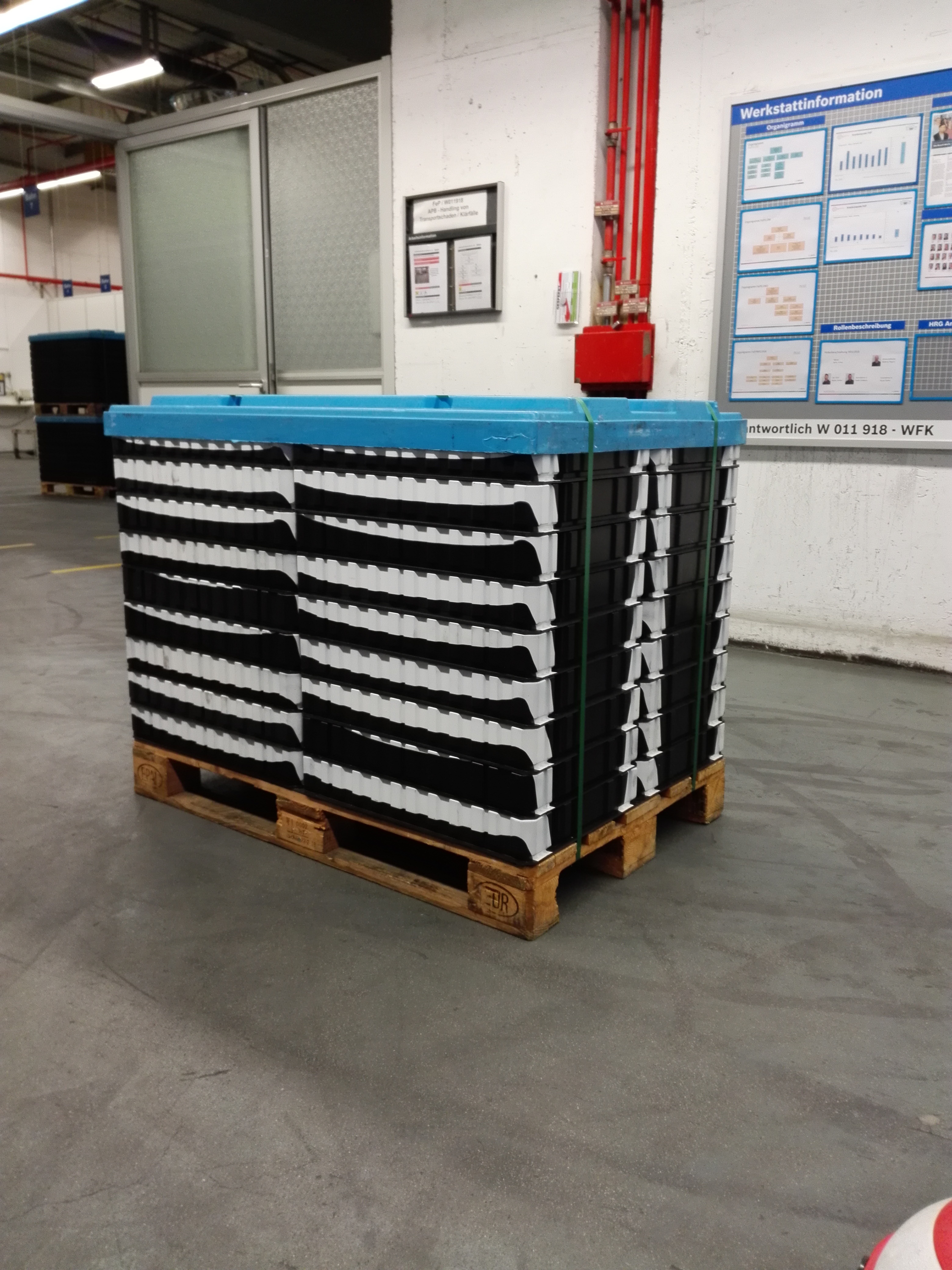}
	\caption{Example images from our data set. The left two images contain transport units with KLT packaging units, the right two images show transport units with tray packaging units.}
	\label{fig:dataset_images}
\end{figure*}

A specific data set of 1267 images was acquired in a German plant of the automotive sector.
All images comply to the restriction and assumptions described in section \ref{sec:setting}.
As relevant for the setting in consideration, two different types of packaging units are present in the images: KLT packages and tray packages.
Each image contains one single or multiple stacked transport units, which were thoroughly annotated, i.e. transport units, side faces, packaging units and base pallet are labeled on pixel basis.
Of these 1267 images, 163 images are marked as dedicated evaluation data. The other images may be used for algorithm development, fitting and training purposes.

\subsection{Transport Unit Segmentation}
\label{sec:exp_step1}

The step of transport unit segmentation is performed using a convolutional neural network (CNN) for instance segmentation.
Namely, a Mask R-CNN \cite{he2017mask} architecture with a Inception-v2 \cite{ioffe2015batch} feature extractor was trained using tensorflow 1.14 and the tensorflow object detection API.
The model was pre-trained on the COCO object detection data set \cite{lin2014coco} and fine-tuned for the single-class task of recognizing and segmenting fully visible logistics transports units.
Evaluation of the CNN and the transport unit segmentation step can be found in \cite{doerr2020psr}.

\subsection{Transport Unit Side Segmentation}

Two different approaches for the task of transport unit side segmentation are considered:
One approach is based on machine learning, the other one employs classic image processing techniques.

\textbf{First Approach: CNN.}
The first approaches uses a CNN for instance segmentation of analogous architecture to section \ref{sec:exp_step1}.
Using $75\%$ the dataset's 1104 labeled training and validation images, the model was trained to recognize transport unit side faces, which are thoroughly labeled by four corner points each.
The model achieved an mean average precision (mAP) of 0.877 on validation data ($25\%$ of the training images which were not used in model training) and 0.892 on the 163 evaluation images. 
Hereby, the mean average precision was computed in accordance to the COCO object detection's metric, i.e. as the averaged precision values at different intersection over union (IoU) thresholds of 0.5 to 0.95.
To achieve the desired output format, a post-processing step, which fits four corner points to the instance segmentation mask found by the CNN model, is used.
Hereby, an optimization problem choosing pixel coordinates for the corner points maximizing the region overlap with the CNN output mask, is solved.
For more details, see \cite{doerr2020psr}.

%

\begin{figure}[htb]
	\centering
	\includegraphics[height=0.2\textheight]{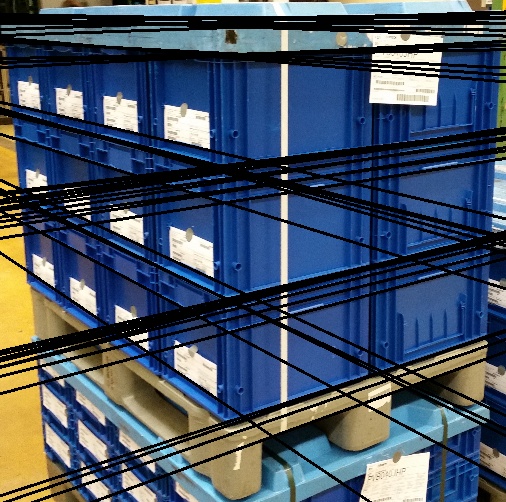}
	\includegraphics[height=0.2\textheight]{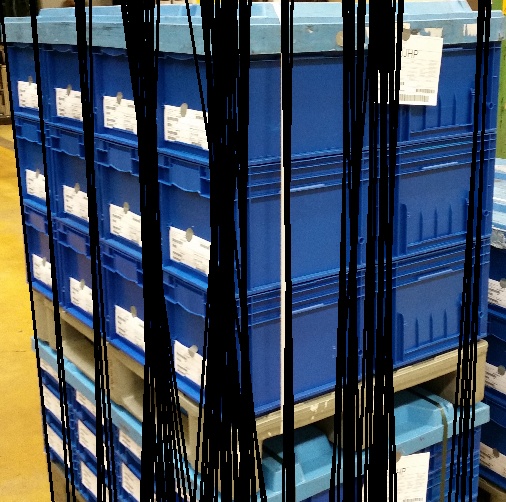}
	\hspace{0.12\textwidth}\includegraphics[height=0.2\textheight]{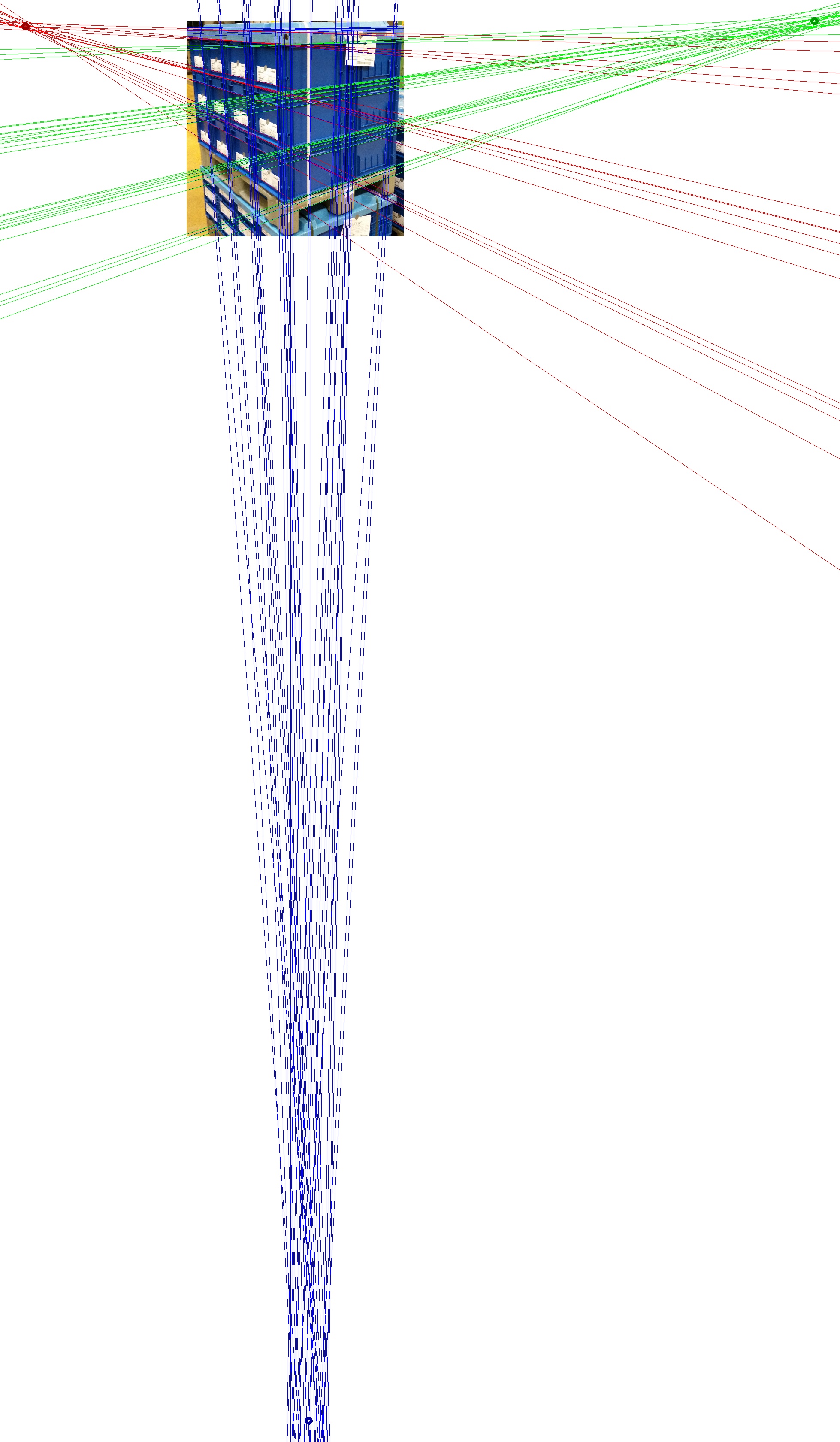}\hspace{0.17\textwidth} \\
	(a) \hspace{0.28\textwidth} (b) \hspace{0.28\textwidth} (c) \\ \vspace{0.1cm}
	\includegraphics[height=0.2\textheight]{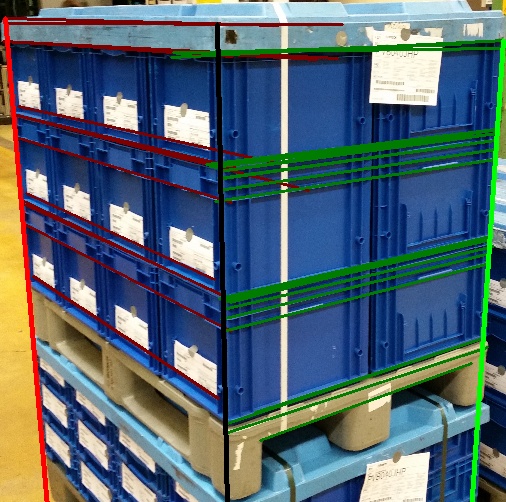}
	\includegraphics[height=0.2\textheight]{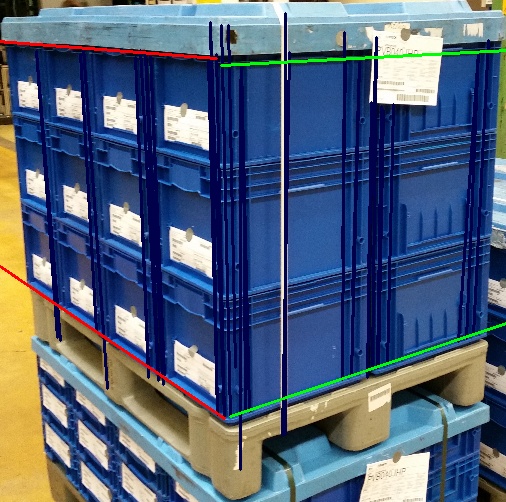}
	\includegraphics[height=0.2\textheight]{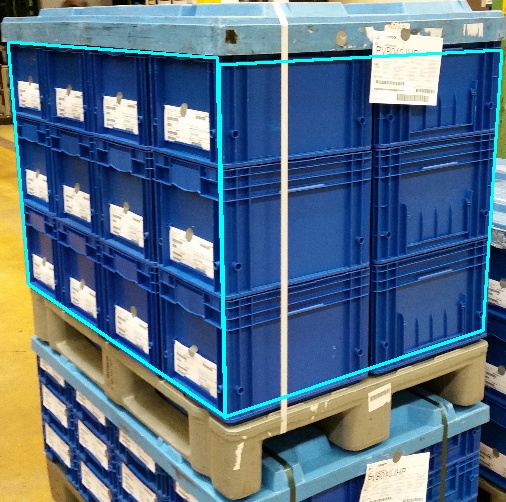} \\
	(d) \hspace{0.28\textwidth} (e) \hspace{0.28\textwidth} (f) 
	\caption{Segmentation of transport unit sides. Detected (a) horizontal and (b) vertical line segments, and (c) vanishing points of transport unit sides. (d), (e) Determination of boundary lines. (f) Resulting transport unit sides.}
	\label{fig:side_seg_hough}
\end{figure}

\textbf{Second Approach: Image Processing.}
Secondly, an image processing approach based on the Hough transform \cite{hough1962method}, a well-established method for detecting straight lines in images, was implemented.
As package and transport units are of regular, rectangular shapes, an image crop showing one transport unit contains many linear structures.
The approach's objective is to detect these linear structures, especially the edges of packaging units, to determine the transport unit side regions within the image.
This is done in the following steps:

\textit{1. Line Detection:}
To detect qualifying horizontal and vertical structures, two different edge detection filter kernels are applied to the image, and the resulting edge images are binarized.
Thereby, the image foreground is restricted to the actual transport unit region using the pixel mask which is input to the step of transport unit side segmentation.
The binary images are used as input for the Hough transform in order to find linear structures.
The line detection results are illustrated in Fig. \ref{fig:side_seg_hough} (a) and (b).

\textit{2. Vanishing Point Estimation:}
After the line detection has been performed, we try to determine the image's vanishing points \cite{barnard1983vanishingpoints} for vertical lines and for the visible transport unit sides' horizontal lines.
To do so, we use a heuristic approach exploiting the knowledge on the image's contents and its geometric properties.
We assume that the majority of vertical line segments detected correspond to vertical edges of the transport unit.
After computing all intersection points of these vertical lines, we use the mean value of all intersection points as first guess for the unit's vertical vanishing point.
Iteratively, we drop lines which do not get sufficiently close to the current vanishing point estimate.
Then, we refine the estimate based on the intersection points of the reduced set of lines. 
This step is repeated several times with decreasing distance thresholds to obtain the final vanishing point position.
For the two vanishing points of horizontal lines on our transport unit sides, we proceed similarly.
We first try to find two accumulation points of horizontal line intersections: One on the left-hand side of the image and one on the right-hand side.
Once again, we repeatedly assign lines in the vicinity of the vanishing points to its line set and use the reduced sets of lines to refine the vanishing point positions.
Fig. \ref{fig:side_seg_hough} (c) illustrates vanishing point estimation and line assignments.

\textit{3. Side Boundary Estimation:}
Based on the vanishing points and corresponding lines, we try to segment the transport unit sides.
To do so, start and end points for all horizontal line segments are determined by matching the line coordinates back to the binary edge image which we used before as input to the Hough operator.
Using the obtained line endpoints, we estimate the transport unit side boundaries by fitting regression lines through corresponding endpoints of each line set and the vanishing point of the side's orthogonal lines.
For instance, to find the left boundary of the left transport unit side, we regress a line through the vertical vanishing point and the top-most endpoints of all lines assigned to that vanishing point.
The transport unit side corner points are inferred by intersecting these boundary lines.
This is illustrated in Fig. \ref{fig:side_seg_hough} (d)-(f).

Overall, there is a considerable number of thresholds and similar parameters contained in this approach. 
For instance, in finding binary edge images, parameters involved are kernels sizes and patterns and binarization threshold.
Further parameters are required when performing the Hough transform, e.g. the minimum length of line segments to consider, as well as distance and angle resolutions.
Also, in vanishing point estimation and line assignments, and in line endpoint determination, numerous threshold parameters are involved.

\textbf{Evaluation.}
To evaluate the complete task of transport unit side segmentation, two different values are considered.
First of all, the average intersection over union (IoU) for all transport unit sides is computed.
Additionally, assuming sufficient accuracy to be given at an IoU of at least 0.8, the number of transport unit sides detected correctly is calculated.
The results for both methods in consideration are shown in Table \ref{tab:side_seg_results}.
The values show that, in the current implementation, the CNN outperforms our image processing approach by a great margin.

\begin{table}
	\centering
	\begin{tabular}{|l|c|c|}
		\hline
		Method & Average IoU & Accuracy \\
		\hline
		CNN & 0.8962 & 0.9029 \\
		Image Processing & 0.6346 & 0.3006 \\
		\hline
	\end{tabular}
	\caption{Transport unit side segmentation evaluation results.}
	\label{tab:side_seg_results}
\end{table}

Even though it is possible to tune the image processing algorithm to deliver precise results for single images or groups of images, we were not successful in finding parameters yielding good results on the whole data set.
The evaluation values shown are the best values achieved by systematically varying the involved parameters in grid-search-like fashion.
The CNN, on the other hand, easily generalizes to data as diverse as ours, due to the huge number of learnable parameters.
Thus, the learning based algorithm appears superior, if not willing to distinguish different groups of images (e.g. by packaging type or size).

\subsection{Packaging Unit Segmentation}

For packaging unit segmentation, a CNN model analogously to section \ref{sec:exp_step1}, is used.
The model performs significantly better on KLT units compared to tray units, which is visible in the per class precision values (0.76 for KLT units compared to 0.67 for tray units), and in the overall error values for the different packaging types (see \cite{doerr2020psr} for details).

First experiments applying image processing operations suggest that their application might be beneficial, especially in the case of tray packaging units.
Package number determination could be tackled by analysing distances and frequencies of detected line segments in a rectified version of the transport unit side image.
We plan to investigate this and conduct detailed experiments on that behalf in the future.

\subsection{Pipeline Evaluation}

In an end-to-end evaluation, the CNN-based packaging structure recognition pipeline achieved an overall accuracy value of approximately $84\%$. 
The metric applied was a custom, use-case specific metric measuring the average ratio of correctly recognized and analyzed transport units per image.
Again, more details can be found in \cite{doerr2020psr}.

\section{Summary}

We presented the problem of packaging structure recognition from single RGB images.
For a specific logistics setting, we formulated reasonable restrictions and assumptions, and designed and presented a solution approach for this setting.
The multi-step image processing pipeline was discussed and evaluated step by step, on our own use-case specific data set.
Specifically for the step of transport unit side recognition, two different algorithms were implemented and compared systematically: 
A learning-based CNN for instance segmentation and a classic computer vision approach based on edge detection.
The first outperformed the latter by a significant margin, which can be accounted for by the high variance in our image set and the CNN's superior generalization ability due to the higher number of parameters.

	\bibliography{bib}

\begin{thebibliography}{10}

\bibitem{barnard1983vanishingpoints}
S.~T. Barnard.
\newblock Interpreting perspective images.
\newblock {\em Artificial intelligence}, 21(4):435--462, 1983.

\bibitem{Solution:cognex}
Cognex.
\newblock Logistics {Industry} {Solutions}.
\newblock \url{https://www.cognex.com/industries/logistics}.
\newblock Accessed: 2020-09-28.

\bibitem{doerr2020psr}
L.~D{\"o}rr, F.~Brandt, M.~Pouls, and A.~Naumann.
\newblock Fully-automated packaging structure recognition in logistics
  environments.
\newblock {\em arXiv preprint arXiv:2008.04620}, 2020.

\bibitem{EPAL:pallet}
{European Pallet Association e.V. (EPAL)}.
\newblock {EPAL Euro Pallet (EPAL 1)}.
\newblock
  \url{https://www.epal-pallets.org/eu-en/load-carriers/epal-euro-pallet/}.
\newblock Accessed: 2020-09-28.

\bibitem{VDA:KLT}
{German Association of the Automotive Industry}.
\newblock {VDA 4500 - Small Load Carrier (SLC) System (KLT)}.
\newblock
  \url{https://www.vda.de/en/services/Publications/small-load-carrier-(slc)-system-(klt).html}.
\newblock Accessed: 2020-09-28.

\bibitem{he2017mask}
K.~He, G.~Gkioxari, P.~Doll{\'a}r, and R.~Girshick.
\newblock Mask r-cnn.
\newblock In {\em Proceedings of the IEEE international conference on computer
  vision}, pages 2961--2969, 2017.

\bibitem{Solution:Fraunhofer_Ladungstraeger}
J.~Hinxlage and J.~Möller.
\newblock Ladungsträgerzahlung per smartphone.
\newblock {\em Jahresbericht Fraunhofer IML 2018}, pages 72--73, 2018.

\bibitem{hough1962method}
P.~V. Hough.
\newblock Method and means for recognizing complex patterns, Dec.~18 1962.
\newblock US Patent 3,069,654.

\bibitem{ioffe2015batch}
S.~Ioffe and C.~Szegedy.
\newblock Batch normalization: Accelerating deep network training by reducing
  internal covariate shift.
\newblock {\em arXiv preprint arXiv:1502.03167}, 2015.

\bibitem{lim2013rfid}
M.~K. Lim, W.~Bahr, and S.~C. Leung.
\newblock Rfid in the warehouse: A literature analysis (1995--2010) of its
  applications, benefits, challenges and future trends.
\newblock {\em International Journal of Production Economics}, 145(1):409--430,
  2013.

\bibitem{lin2014coco}
T.-Y. Lin, M.~Maire, S.~Belongie, J.~Hays, P.~Perona, D.~Ramanan,
  P.~Doll{\'a}r, and C.~L. Zitnick.
\newblock Microsoft coco: Common objects in context.
\newblock In {\em European conference on computer vision}, pages 740--755.
  Springer, 2014.

\bibitem{Solution:Logivations}
Logivations.
\newblock Ai-based identification solutions in logistics.
\newblock
  \url{https://www.logivations.com/en/solutions/agv/camera_identification.php}.
\newblock Accessed: 2020-09-28.

\bibitem{Solution:vitronic}
Vitronic.
\newblock Warehouse \& distribution logistics.
\newblock
  \url{https://www.vitronic.com/en-us/logistics/warehouse-and-distribution}.
\newblock Accessed: 2020-09-28.

\end{thebibliography}
	

	\renewcommand{\bibname}{References}
	\cleardoublepage

\end{document}